\pdfoutput=1

\documentclass[11pt]{article}

\usepackage[]{EMNLP2023}

\usepackage{times}
\usepackage{latexsym}
\usepackage{hyperref}
\usepackage{color, colortbl}
\definecolor{Gray}{gray}{0.9}
\usepackage{array}
\usepackage{booktabs}
\usepackage{multirow}
\usepackage{amsmath}
\usepackage{amssymb}
\usepackage{graphicx}
\usepackage{enumerate}
\usepackage{diagbox}
\usepackage{mathtools}
\usepackage{paralist}
\usepackage{tabularx}
\usepackage{lipsum}
\usepackage{ragged2e}
\usepackage{float}
\usepackage{tabularx}
\usepackage[capitalize,noabbrev]{cleveref}

\usepackage[T1]{fontenc}

\usepackage[utf8]{inputenc}

\usepackage{microtype}

\newcommand*{\affaddr}[1]{#1} 
\newcommand*{\affmark}[1][*]
{\textsuperscript{#1}}

\usepackage{multirow}
\title{TILFA: A Unified Framework for Text, Image, and Layout Fusion in Argument Mining}

\author{
Qing Zong\affmark[1],
Zhaowei Wang\affmark[2],
Baixuan Xu\affmark[2],
Tianshi Zheng\affmark[2],
Haochen Shi\affmark[2],\\
\textbf{
Weiqi Wang\affmark[2],
Yangqiu Song\affmark[2],
Ginny Y. Wong\affmark[3], 
Simon See\affmark[3]}\\
\affaddr{\affmark[1]Harbin Institute of Technology~(Shenzhen), Guangdong, China} \\
\affaddr{\affmark[2]Department of Computer Science and Engineering, HKUST, Hong Kong SAR, China}\\
\affaddr{\affmark[3]NVIDIA AI Technology Center (NVAITC), NVIDIA, Santa Clara, USA}\\
\texttt{zongqing0068@gmail.com,
\{bxuan, tzhengad, hshiah\}@connect.ust.hk,}\\ 
\texttt{\{zwanggy, wwangbw, yqsong\}@cse.ust.hk}}

\begin{document}
\maketitle
\begin{abstract}

A main goal of Argument Mining (AM) is to analyze an author's stance. Unlike previous AM datasets focusing only on text, the shared task at the 10\emph{th} Workshop on Argument Mining introduces a dataset including both text and images. Importantly, these images contain both visual elements and optical characters. Our new framework, \textbf{TILFA}\footnote{The code and data are available at \url{https://github.com/HKUST-KnowComp/TILFA}.}~(A Unified Framework for \textbf{T}ext, \textbf{I}mage, and \textbf{L}ayout \textbf{F}usion in \textbf{A}rgument Mining), is designed to handle this mixed data. It excels at not only understanding text but also detecting optical characters and recognizing layout details in images.
Our model significantly outperforms existing baselines, earning our team, \textbf{KnowComp}, the \textbf{1\emph{st}} place in the leaderboard\footnote{\url{https://imagearg.github.io/}} of Argumentative Stance Classification subtask in this shared task.

\end{abstract}

\section{Introduction}
Argument mining~(AM) aims to automatically analyze the structure and components of arguments in text. Persuasiveness analysis is a crucial aspect of it, which has gained significant attention in the NLP community~\citep{
habernal-gurevych-2017-argumentation, 
carlile-etal-2018-give}. However, previous works focus solely on text, overlooking other modalities like images which can also impact an argument's persuasiveness.
To fill this gap, \citet{liu-etal-2022-imagearg} introduces \textbf{ImageArg}, a dataset going beyond text to include also images. It features tweets centered on contentious topics like gun control and abortion. These associated images contain not only objects but also optical characters (e.g., slogans, tables).

\begin{figure}[ht]
\centering
\includegraphics[scale=0.35]{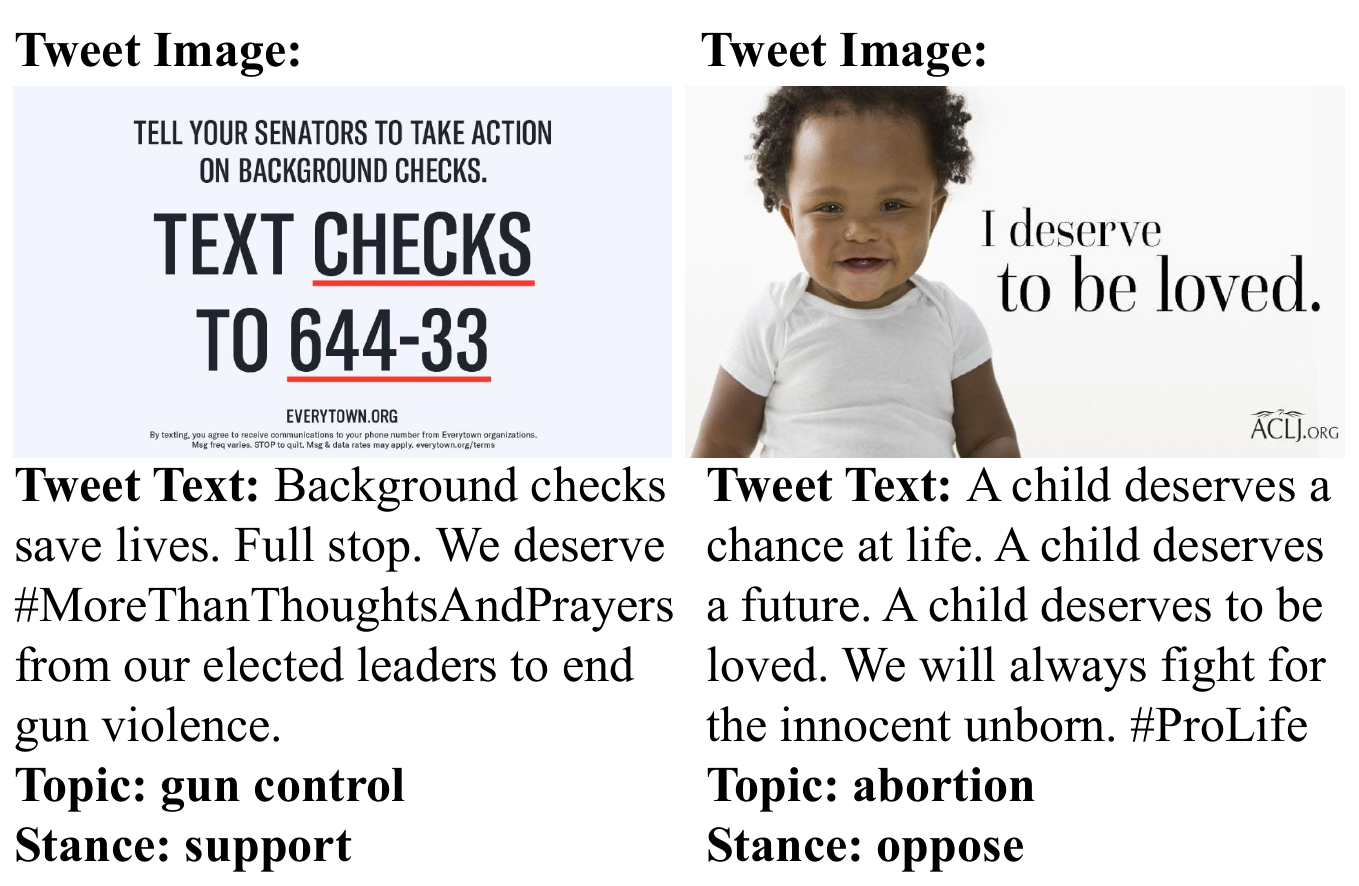}
\caption{Examples of positive (support) and negative (oppose) tweets of different topics. The images also contain a lot of information crucial to stance identification.} 
\label{dataset_image}
\end{figure} 

The 10\emph{th} Workshop on Argument Mining in EMNLP 2023 introduces a shared task~\citep{liu-etal-2023-overview} called ImageArg Shared Task 2023, centering around this dataset. It is divided into two subtasks: Argumentative Stance (AS) Classification and Image Persuasiveness (IP) Classification. We primarily focus on the former, which aims to identify the stance of a given tweet towards a specific topic. Examples can be found in Fig.~\ref{dataset_image}. 

After scrutinizing the dataset, we found several challenges: (1) Imbalanced data distribution~(Ratio of positive to negative examples on abortion topic is about $1:2.65$); (2) Limited data size~(Neither of the two topics has more than 1000 entries); (3) Presence of both objects and optical characters in images~(They are difficult to be handled by a single model at the same time). To address these challenges, we have made the following contributions:

\begin{itemize}
    \item To tackle data imbalance, we employ back-translation to enrich data in the fewer class, as described by \citet{QANet, wieting2018paranmt}.
    \item For data augmentation, we utilize WordNet~\citep{miller-1994-wordnet} with GlossBERT~\citep{huang-etal-2019-glossbert} to create additional data by replacing synonyms of nouns in original instances.
    \item We introduce \textbf{TILFA} which can understand both text and image well, especially adept at detecting optical characters and discerning layout details in images.
\end{itemize}

\section{Related work}

\paragraph{Data Augmentation: } Data augmentation enhances a model’s performance and increases its generalization capabilities in Natural Language Processing (NLP). 
At the word level, \citet{
Wang} used databases like WordNet~\citep{miller-1994-wordnet}, to replace certain words with their synonyms, while \citet{
Rizos} implemented embedding replacement to find contextually fitting words; 
At the document level, \citet{
QANet} used back-translation, translating data to another language and then back to the source one. 

\begin{figure*}[t]
\centering
\includegraphics[scale=0.43]{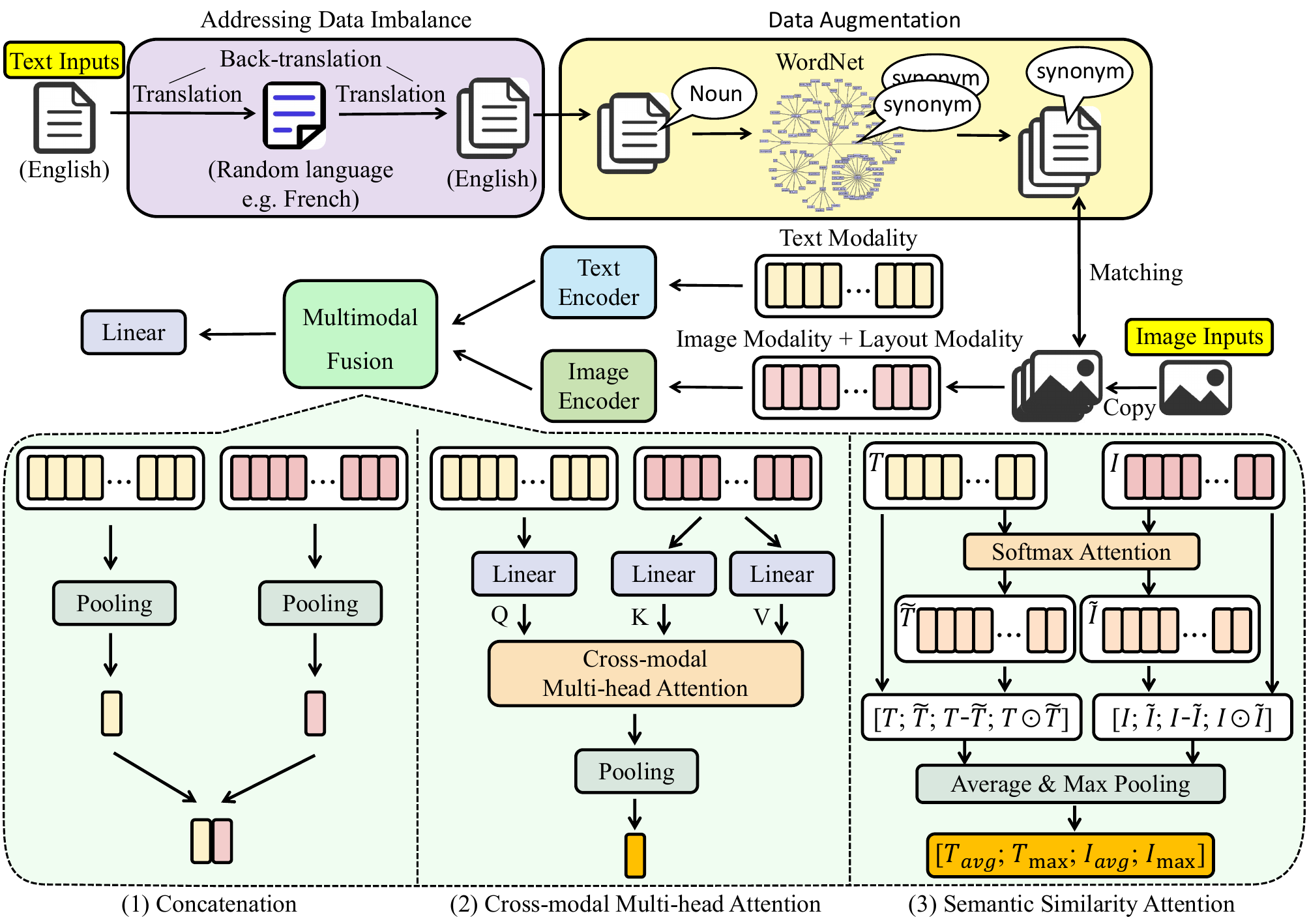} 
\caption{Our model, \textbf{TILFA}, includes three main parts: a text encoder, an image encoder, and a multimodal fusion module. In this fusion module, we experiment with three different methods: (1) Concatenation; (2) Cross-modal Multi-head Attention; (3) Semantic Similarity Attention.} 
\label{model}
\end{figure*} 

\paragraph{Document AI: } Document AI refers to the extraction and comprehension of information from scanned documents, web pages, ads, posters, or images with textual content. 
Previous works like \citet{7, 15} all missed 
the integrated pre-training of text and layout details, which are vital for document image comprehension. To fix this, \citet{xu2019layoutlm} proposed LayoutLM. Its updated versions, LayoutLMv2~\citep{layoutlmv2} and LayoutLMv3~\citep{huang2022layoutlmv3}, encapsulated text, layout and also image interactions within a unified multimodal framework.


\section{Methods}

We employ back-translation~\citep{QANet} to address data imbalance and apply WordNet~\citep{miller-1994-wordnet} for data augmentation, assisted by GlossBERT~\citep{huang-etal-2019-glossbert}. Our new framework, \textbf{TILFA}, uses DeBERTa~\citep{he2021deberta} as the text encoder and LayoutLMv3~\citep{huang2022layoutlmv3} as the image encoder, thus excels at not only understanding text but also detecting optical characters and recognizing layout details in images. We also experiment with several multimodal fusion mechanisms. Consequently, we achieve the \textbf{highest} F1-score in the Argumentative Stance Classification subtask of the ImageArg Shared Task 2023.

\subsection{Addressing Data Imbalance}

Label imbalance is serious in the training set, particularly concerning the abortion topic~(The ratio of positive to negative examples is about $1:2.65$). To address this, we preprocess the data through bask-translation~\citep{QANet}. We translate the English tweet text belonging to the underrepresented label (e.g., positive in abortion topic) to a random language (e.g., French, German) and then back to English. This maintains the tweet's meaning and thus the stance label. The translated text is finally paired with its original image.

\subsection{Data Augmentation}

More data usually leads to better model performance~\citep{AugmentationSurvey,DBLP:journals/corr/abs-2212-10558}. We employ data augmentation methods since our ImageArg training set is limited: only 918 entries for gun control and 888 for abortion. We first utilize spaCy to tokenize the tweet text and extract the nouns in it. Then we find all their synonym sets in WordNet~\citep{miller-1994-wordnet}. We determine these nouns' meanings in context by Word Sense Disambiguation~(WSD) using GlossBERT~\citep{huang-etal-2019-glossbert}, thus getting their correct synonym set. Finally, we replace these nouns with their synonyms to create new data. 

\subsection{Model}
\label{method_model}

To solve this task, we introduce a model:  \textbf{TILFA} (A Unified Framework for \textbf{T}ext, \textbf{I}mage, and \textbf{L}ayout \textbf{F}usion in \textbf{A}rgument Mining). The structure of \textbf{TILFA} is illustrated in Fig.~\ref{model}, comprising three components: Image Encoder, Text Encoder, and Multimodal Fusion. We will discuss details of them one by one.

\paragraph{Image Encoder: }As highlighted by \citet{liu-etal-2022-imagearg}, traditional image encoders like ResNet50, ResNet101~\citep{He_2016_CVPR}, VGG~\citep{VGG} are good at identifying objects but fall short in recognizing optical characters in images, which may hurt performance. However, as shown in Fig.~\ref{dataset_image}, many of the images in our dataset contain significant amount of characters. So we reasonably believe that using models which can capture the text in the images will have better results. And we notice that those characters with more prominent position and larger size are relatively more important. So considering the importance of this layout information of characters in images, we employ LayoutLMv3~\citep{
huang2022layoutlmv3}) to encode the images, favoring it over sole OCR tools. 

\paragraph{Text Encoder: }To encode tweet texts, we employ DeBERTa~\citep{he2021deberta}. It has shown great performance in various NLP tasks . Our experimental results confirm its effectiveness in the ImageArg task as well. 

\paragraph{Multimodal Fusion: }We explore three multimodal fusion methods, illustrated in Fig.~\ref{model}. The first simply concatenates the hidden states from text and image inputs. The second method, named cross-modal multi-head attention, is adapted from \citet{yu-etal-2021-vision}. And the third is a new approach adapted from ESIM~\citep{esim}. ESIM is a sequential natural language inference model used to predict the logic relationship between two sentences. We adapt it for text and image fusion, and name our version: Semantic Similarity Attention.

\section{Experiments}

\subsection{Experiment Settings}

\paragraph{Metrics:} We use F1 score, Macro F1, AUC (Area Under Curve) and accuracy scores to evaluate baselines and our models. Models with the best F1-score on validation set are chosen.

\paragraph{Baselines:} For images, we use ResNet50, ResNet101, VGG16, and LayoutLMv3. For text, we use DeBERTa. For combined image and text input, DeBERTa serves as text encoder, while ResNet50, ResNet101, and VGG16 act as image encoder. All models use the original dataset.

\paragraph{Implementation Details: }
To be more specific, we report scores within topics.
Since the hyperparameters have a non-negligible effect on the scores, we conduct the experiments at a learning rate of 1e-4, 1e-5, 5e-6 and a batch size of 16, 8, 4. More implementation details can be found in Appendix~\ref{Implementation Details}.

\begin{table*}[t]
    \resizebox{\linewidth}{!}{
        \centering
	\begin{tabular}{l|l||cccc|cccc}
	\toprule
        \multirow{2}{*}{\textbf{Models}}&\multirow{2}{*}{\textbf{Modality}}&\multicolumn{4}{c|}{\textbf{Gun Control}} &\multicolumn{4}{c}{\textbf{Abortion}}\\
	&&\textbf{F1} &\textbf{Ma-F1} &\textbf{AUC} &\textbf{Acc} &\textbf{F1} &\textbf{Ma-F1} & \textbf{AUC} &\textbf{Acc} \\
        \midrule
        ResNet50 &I& 65.00 & 62.50 & 62.90 & 62.67 & 35.71 & 60.42 & 59.55 & 75.84\\
        ResNet101 &I& 67.43 & 60.91 & 60.86 & 62.00 & 38.89 & 59.71 & 60.44 & 70.47\\
        VGG16 &I& 67.44 & 61.85 & 61.81 & 62.67 & 38.46 & 58.32 & 59.80 & 67.79 \\
        LayoutLMv3 & I+L & 56.98 & 46.67 & 46.92 & 48.67 & 39.47 & 59.38 & 60.66 & 69.13 \\
        DeBERTa &T& 86.32 & 81.34 & 80.54 & 82.67 & 71.43 & 80.11 & 86.40 & 83.89 \\
        DeBERTa+ResNet50 &T+I& \textbf{88.04} & \textbf{84.54} & \textbf{83.80} & \textbf{85.33} & 70.59 & 79.43 & 85.97 & 83.22\\
        DeBERTa+ResNet101 &T+I& 87.50 & 82.64 & 81.72 & 84.00 & 73.17 & 81.49 & 87.26 & 85.23\\
        DeBERTa+VGG16 &T+I& 87.96 & 83.43 & 82.49 & 84.67 & \textbf{74.07} & \textbf{82.20} & \textbf{87.70} & \textbf{85.91} \\
        \midrule
        \midrule
        DeBERTa+ResNet50 &T+I $^+$ & \textbf{90.32} & \textbf{87.27} & \textbf{86.33} & \textbf{88.00} & 75.95 & 83.64 & 88.56 & 87.25\\
        DeBERTa+ResNet101 &T+I $^+$ & 87.43 & 83.89 & 83.21 & 84.67 & 70.73 & 79.81 & 85.32 & 83.89\\
        DeBERTa+VGG16 &T+I $^+$ & 89.73 & 86.60 & 85.75 & 87.33 & \textbf{77.50} & \textbf{84.62} & \textbf{90.07} & \textbf{87.92} \\
        \midrule
        TILFA~(DeBERTa+LayoutLMv3) & t1 T+I+L & 89.13 & 85.94 & 85.16 & 86.67 & 76.54 & 83.89 & 89.64 & 87.25\\
        TILFA~(DeBERTa+LayoutLMv3) & t1 T+I+L $^+$ & \textbf{90.81} & \textbf{88.01} & \textbf{87.10} & \textbf{88.67} & \textbf{80.52} & \textbf{86.87} & 91.37 & \textbf{89.93}\\
        TILFA~(DeBERTa+LayoutLMv3) & t2 T+I+L $^+$ & 90.32 & 87.27 & 86.33 & 88.00 & 77.50 & 84.62 & 90.07 & 87.92\\
        TILFA~(DeBERTa+LayoutLMv3) & t3 T+I+L $^+$ & 88.89 & 84.98 & 84.03 & 86.00 & 79.01 & 85.59 & \textbf{91.59} & 88.59\\
        \bottomrule
        \end{tabular}
    }
        \caption{Performance of all frameworks on the testing set of both topics. Those below the double horizontal line use our methods, and the above are baselines. For models that have both base and large sizes, we use the large one. We abbreviate F1-score, Macro F1-score, Accuracy to F1, Ma-F1, Acc, respectively. T, I and L are short for text modality, image modality and layout modality. Three multimodal fusion methods are named t1, t2 and t3 here. Those with a superscript $+$ use both back-translation and WordNet, while others don't use either. }
	\label{part_of_scores}
\end{table*}

\subsection{Results and Analysis}

\cref{part_of_scores} shows the results on both topics in ImageArg dataset. (Experimental results of more models can be seen in Appendix~\ref{Experimental Results}.)

Our model, \textbf{TILFA}, outperforms the baselines~(those above the double horizontal line) on all the four evaluation metrics by a large margin. Also, our model achieves the SOTA performance on the leaderboard of Argumentative Stance Classification subtask in ImageArg Shared Task 2023, which demonstrate the effectiveness of our methods.

For combined text and image inputs, models utilizing LayoutLMv3 for image encoding perform much better than those using traditional ones, which verifies our belief in Section~\ref{method_model} that a better understanding of the text in images is beneficial. 

Back-translation and WordNet also greatly improve performance across all metrics~(e.g. an improvement of 1.68 on gun control and 3.98 on abortion for \textbf{TILFA} in F1-score), confirming the value of our data preprocessing and augmentation strategies. When it comes to multimodal fusion methods, the simplest Concatenation works best. We think it may because the second method
is initially applied in video field~\citep{yu-etal-2021-vision}, and the third one
in pure text field~\citep{esim}. So, neither of them is suitable to be migrated to this task.

We merge the answers belonging to different topics together and report the Micro F1-score. With a 90.32 F1 on gun control and a 77.50 F1 on abortion, we get a Micro F1-score of 86.47, which is the top on the leaderboard, 1.41 higher than the second best team. Our score improves even further after changing the hyperparameters, up to 87.79~(90.81 on gun control and 80.52 on abortion).

\subsection{Ablation Study}

To fully understand the impact of different components, we conduct an ablation study 
in \cref{Ablation_study}.

\begin{table}[t]
    \resizebox{\linewidth}{!}{
	\centering
	\begin{tabular}{c|c|c|c|cc}
	\toprule
        \multirow{2}{*}{\textbf{Text}}&\multirow{2}{*}{\textbf{Image}}&\multirow{2}{*}{\textbf{T}}&\multirow{2}{*}{\textbf{W}}&\multicolumn{2}{c}{\textbf{F1}}\\
	&&&&\textbf{Gun} &\textbf{Abortion} \\
        \midrule
        $-$ & ResNet50 & $-$ & $-$ & 65.00 & 35.71 \\
        $-$ & LayoutLMv3 & $-$ & $-$ & 56.98 & 39.47 \\
        DeBERTa & $-$ & $-$ & $-$ & 86.32 & 71.43 \\
        \midrule
        DeBERTa & ResNet50 & $-$ & $-$ & 88.04 & 70.59 \\
        DeBERTa & ResNet50 & \checkmark & $-$ & 88.42 & 71.11 \\
        DeBERTa & ResNet50 & $-$ & \checkmark & 88.77 & 75.61 \\
        DeBERTa & ResNet50 & \checkmark & \checkmark & 90.32 & 75.95 \\
        \midrule
        DeBERTa & LayoutLMv3 & $-$ & $-$ & 89.13 & 76.54 \\
        DeBERTa & LayoutLMv3 & \checkmark & $-$ & 89.73 & 76.92 \\
        DeBERTa & LayoutLMv3 & $-$ & \checkmark & 90.22 & 77.50 \\
        DeBERTa & LayoutLMv3 & \checkmark & \checkmark & 90.81 & 80.52 \\
        \bottomrule
        \end{tabular}
    }
        \caption{Ablation studies in ImageArg. The first two columns illustrate text and image encoders. T and W represent back-translation and WordNet. All models use the Concatenation method for multimodal fusion.}
    \label{Ablation_study}
\end{table}

Both back-translation and WordNet do help to the improvement of model performance, with WordNet having a larger impact. Models using only text inputs outperform just image inputs. This suggests that information in the text is more effective in determining the author's stance than images. However, the best performance is achieved when both text and image inputs are used, showing that images also do contribute to stance determination. 

LayoutLMv3 performs better than Resnet50 on abortion topic when based solely on image inputs, but on both topics when text inputs are added. This indicates that image encoders which can take the text and layout information in the images into account can really work better.

\subsection{Case Study}

\begin{table*}[!htbp]
	\centering
	\begin{tabularx}{\textwidth}{lX}
		\toprule
		Original Text 
		& SCOTUS has balanced rights w/ public safety, ruling that gun safety laws essential \& constitutional Rushing through a replacement to RBG could undermine that balance and put life-saving laws at risk. \\
        \midrule
        Selected Noun & "risk" \\
        \midrule
        Word Senses & 
        "hazard.n.01": a source of danger; a possibility of incurring loss or misfortune

        "risk.n.02": a venture undertaken without regard to possible loss or injury
        
        "risk.n.03": the probability of becoming infected given that exposure to an infectious agent has occurred
        
        "risk.n.04": the probability of being exposed to an infectious agent
        
        "risk.v.01": expose to a chance of loss or damage
        
        "gamble.v.01": take a risk in the hope of a favorable outcome	
        \\
        \midrule
        Disambiguation & "hazard.n.01": a source of danger; a possibility of incurring loss or misfortune \\
        \midrule
        Synonyms & "hazard.n.01" = ["peril", "jeopardy", "endangerment", "hazard"]\\
        \midrule
        New Text & SCOTUS has balanced rights w/ public safety, ruling that gun safety laws essential \& constitutional Rushing through a replacement to RBG could undermine that balance and put life-saving laws at peril/jeopardy/endangerment/hazard. \\
        \bottomrule
	\end{tabularx}
	\caption{An example of our data augmentation method.}
	\label{case_study}
 \vspace{-0.1in}
\end{table*}

We conduct a case study to better understand the behavior of our data augmentation method, with an example presented in \cref{case_study}. In the original text, we select the noun "risk". Then we find its different meanings and corresponding synonym sets in WordNet. Using GlossBERT, we determine its exact meaning "hazard.n.01" and thus get the correct synonym set ["peril", "jeopardy", "endangerment", "hazard"]. Finally, we replace the noun "risk" in the original text with these synonyms to form new text.

\section{Conclusion}

We present \textbf{TILFA}, a new framework for multimodal argumentative stance classification. Unlike existing methods, \textbf{TILFA} considers not only the text and images in tweets but also the characters and their layout information in those images. Back-translation and WordNet also contribute to our SOTA performance. Our results reveal that better handling of images is essential to model improvement, and suggest that more effective methods for multimodal fusion are yet to be found.

\section*{Limitations}

We have experimented with three multimodal fusion methods, but the simplest one, Concatenation, turned out to be the best. So the other two methods that we use are not suitable for this task actually. But we believe that there are more effective multimodal fusion methods~\cite{liu2023incomplete, li1, yang1} waiting to be discovered. 

Also, we notice that images in the dataset vary widely, some feature only objects, but others contain significant text. Further research is needed to effectively handle these differences, and we expect that better image encoders will improve performance in future works. 

Moerover, in the data augmentation part, we only explore the methods related to text, but there are also many ways to augment images. Whether these methods~\cite{survey_image, XU2023109347} are effective for images containing lots of characters is a question worth studying.

For the text modality, we found that most instances are a piece of text containing a few events, such as the example in \cref{case_study}. With the recent advances in event understanding~\cite{lin2023global}, we can incorporate different relations among events, including temporal~\cite{fang2023getting}, causal~\cite{zhang2022rock, DBLP:conf/acl/0003DZ0WFSWS23, gao2023chatgpt}, sub-event~\cite{DBLP:conf/emnlp/WangZFSWS22, zhang2020analogous}, hierarchical~\cite{wang-etal-2023-cat, DBLP:journals/corr/abs-2305-14869}. 

\section*{Acknowledgements}
The authors of this paper were supported by the NSFC Fund (U20B2053) from the NSFC of China, the RIF (R6020-19 and R6021-20), and the GRF (16211520 and 16205322) from RGC of Hong Kong. We also thank the support from NVIDIA AI Technology Center (NVAITC) and the UGC Research Matching Grants (RMGS20EG01-D, RMGS20CR11, RMGS20CR12, RMGS20EG19, RMGS20EG21, RMGS23CR05, RMGS23EG08).


\bibliography{emnlp2023-latex/emnlp2023}
\bibliographystyle{acl_natbib}

\newpage
\appendix

\section{Implementation Details}
\label{Implementation Details}

In this appendix, we introduce the implementation details of every component in our framework \textbf{TILFA}.

We use the dataset divisions provided in the shared task, and the dataset sizes are detailed in \cref{dataset_scale}.
Each tweet text is cut by a maximum length of 512, and each tweet image is resized to $224\times224$ dimension. 
For back-translation, we use Youdao translation API\footnote{\url{http://fanyi.youdao.com/openapi/}}. For layout information, we follow \citet{xu2019layoutlm} and use Tesseract\footnote{\url{https://github.com/tesseract-ocr/tesseract}}, an open-source OCR engine, to get the recognized words and their 2-D positions in the images. 
Our models are implemented with Pytorch, and trained on a NVIDIA A6000 GPU. AdamW optimizer is used for those networks with LayoutLMv3 and Adam optimizer for others.

\begin{table}[h]
    \small
	\centering
	\begin{tabular}{c|c|c|c}
	\toprule
        \textbf{Topic}&\textbf{Train}&\textbf{Validation}&\textbf{Test} \\
        \midrule
        Gun control & 918 & 96 & 150 \\
        Abortion & 888 & 100 & 149 \\
        \bottomrule
        \end{tabular}
        \caption{Dataset scale of both topics. Following the shared task, one unavailable tweet in the abortion testing set is removed. And due to the downloading issues, our downloaded train and dev sets have little difference from the original one.}
    \label{dataset_scale}
\end{table}

\section{Experimental Results}
\label{Experimental Results}
Our full experiment results are shown in Table~\ref{scores}.

\begin{table*}[h]
    \small
	\centering
        \resizebox{\linewidth}{!}{
	\begin{tabular}{l|l||cccc|cccc}
	\toprule
        \multirow{2}{*}{\textbf{Models}}&\multirow{2}{*}{\textbf{Modality}}&\multicolumn{4}{c|}{\textbf{Gun Control}} &\multicolumn{4}{c}{\textbf{Abortion}}\\
	&&\textbf{F1} &\textbf{Ma-F1} &\textbf{AUC} &\textbf{Acc} &\textbf{F1} &\textbf{Ma-F1} & \textbf{AUC} &\textbf{Acc} \\
        \midrule
        ResNet50 &I& 65.00 & 62.50 & 62.90 & 62.67 & 35.71 & 60.42 & 59.55 & 75.84\\
        ResNet101 &I& 67.43 & 60.91 & 60.86 & 62.00 & 38.89 & 59.71 & 60.44 & 70.47\\
        VGG16 &I& 67.44 & 61.85 & 61.81 & 62.67 & 38.46 & 58.32 & 59.80 & 67.79 \\
        LayoutLMv3 & I+L & 56.98 & 46.67 & 46.92 & 48.67 & 39.47 & 59.38 & 60.66 & 69.13 \\
        BERT & T& 81.48 & 74.97 & 74.52 & 76.67 & 64.10 & 75.69 & 79.26 & 81.21\\
        RoBERTa & T& 83.52 & 79.05 & 78.55 & 80.00 & 71.60 & 80.50 & 85.75 & 84.56\\
        DeBERTa &T& 86.32 & 81.34 & 80.54 & 82.67 & 71.43 & 80.11 & 86.40 & 83.89 \\
        BERT+ResNet50 &T+I& 81.97 & 76.88 & 76.43 & 78.00 & 65.06 & 75.79 & 81.00 & 80.54\\
        RoBERTa+ResNet50 &T+I& 84.82 & 79.11 & 78.42 & 80.67 & 73.42 & 81.91 & 86.61 &\textbf{85.91} \\
        DeBERTa+ResNet50 &T+I& \textbf{88.04} & \textbf{84.54} & \textbf{83.80} & \textbf{85.33} & 70.59 & 79.43 & 85.97 & 83.22\\
        BERT+ResNet101 &T+I& 80.63 & 73.34 & 72.99 & 75.33 & 64.52 & 74.21 & 82.52 & 77.85\\
        RoBERTa+ResNet101 &T+I& 83.77 & 77.66 & 77.06 & 79.33 & 65.22 & 74.84 & 82.95 & 78.52\\
        DeBERTa+ResNet101 &T+I& 87.50 & 82.64 & 81.72 & 84.00 & 73.17 & 81.49 & 87.26 & 85.23\\
        BERT+VGG16 &T+I& 79.37 & 72.11 & 71.81 & 74.00 & 68.24 & 77.78 & 84.03 & 81.88\\
        RoBERTa+VGG16 &T+I& 83.52 & 79.05 & 78.55 & 80.00 & 72.50 & 81.20 & 86.18 & 85.23\\
        DeBERTa+VGG16 &T+I& 87.96 & 83.43 & 82.49 & 84.67 & \textbf{74.07} & \textbf{82.20} & \textbf{87.70} & \textbf{85.91} \\
        \midrule
        \midrule
        BERT+ResNet50 &T+I $^+$ & 81.05 & 74.16 & 73.76 & 76.00 & 68.24 & 77.78 & 84.03 & 81.88\\
        RoBERTa+ResNet50 &T+I $^+$ & 84.66 & 79.26 & 78.60 & 80.67 & 71.60 & 80.50 & 85.75 & 84.56 \\
        DeBERTa+ResNet50 &T+I $^+$ & \textbf{90.32} & \textbf{87.27} & \textbf{86.33} & \textbf{88.00} & 75.95 & 83.64 & 88.56 & 87.25\\
        BERT+ResNet101 &T+I $^+$ & 82.87 & 78.41 & 77.96 & 79.33 & 62.50 & 72.34 & 81.23 & 75.84\\
        RoBERTa+ResNet101 &T+I $^+$ & 84.21 & 78.47 & 77.83 & 80.00 & 68.97 & 78.08 & 85.11 & 81.88 \\
        DeBERTa+ResNet101 &T+I $^+$ & 87.43 & 83.89 & 83.21 & 84.67 & 70.73 & 79.81 & 85.32 & 83.89\\
        BERT+VGG16 &T+I $^+$ & 81.03 & 72.89 & 72.62 & 75.33 & 68.97 & 78.08 & 85.11 & 81.88 \\
        RoBERTa+VGG16 &T+I $^+$ & 83.24 & 78.14 & 77.60 & 79.33 & 72.50 & 81.20 & 86.18 & 85.23\\
        DeBERTa+VGG16 &T+I $^+$ & 89.73 & 86.60 & 85.75 & 87.33 & \textbf{77.50} & \textbf{84.62} & \textbf{90.07} & \textbf{87.92} \\
        \midrule
        BERT+LayoutLMv3 & t1 T+I+L & 81.32 & 76.25 & 75.84 & 77.33 & 65.75 & 77.32 & 79.47 & 83.22\\
        RoBERTa+LayoutLMv3 & t1 T+I+L & 85.26 & 79.90 & 79.19 & 81.33 & 75.95 & 83.64 & 88.56 & 87.25 \\
        DeBERTa+LayoutLMv3 & t1 T+I+L & 89.13 & 85.94 & 85.16 & 86.67 & 76.54 & 83.89 & 89.64 & 87.25\\
        BERT+LayoutLMv3 & t1 T+I+L $^+$ & 81.72 & 75.95 & 75.48 & 77.33 & 69.14 & 78.81 & 83.80 & 83.22 \\
        RoBERTa+LayoutLMv3 & t1 T+I+L $^+$ & 86.19 & 82.59 & 82.04 & 83.33 &  76.32 & 84.10 & 87.90 & 87.92 \\
        DeBERTa+LayoutLMv3 & t1 T+I+L $^+$ & \textbf{90.81} & \textbf{88.01} & \textbf{87.10} & \textbf{88.67} & \textbf{80.52} & \textbf{86.87} & 91.37 & \textbf{89.93}\\
        BERT+LayoutLMv3 & t2 T+I+L $^+$ & 83.52 & 79.05 & 78.55 & 80.00 & 69.88 & 79.13 & 84.89 & 83.22 \\
        RoBERTa+LayoutLMv3 & t2 T+I+L $^+$ & 84.95 & 80.19 & 79.55 & 81.33 & 72.29 & 80.80 & 86.83 & 84.56 \\
        DeBERTa+LayoutLMv3 & t2 T+I+L $^+$ & 90.32 & 87.27 & 86.33 & 88.00 & 77.50 & 84.62 & 90.07 & 87.92\\
        BERT+LayoutLMv3 & t3 T+I+L $^+$ & 82.16 & 76.73 & 76.24 & 78.00 & 69.05 & 78.45 & 84.46 & 82.55\\
        RoBERTa+LayoutLMv3 & t3 T+I+L $^+$ & 84.78 & 80.32 & 79.73 & 81.33 & 71.05 & 80.57 & 84.01 & 85.23\\
        DeBERTa+LayoutLMv3 & t3 T+I+L $^+$ & 88.89 & 84.98 & 84.03 & 86.00 & 79.01 & 85.59 & \textbf{91.59} & 88.59\\
        \bottomrule
        \end{tabular}
        }
        \caption{Experimental results of more models on both topics in ImageArg dataset. Compared to \cref{part_of_scores}, the scores of the models which use BERT or RoBERTa as the text encoder are also listed here.}
    \label{scores}
\end{table*}

\end{document}